\documentclass{vgtc}                          

\graphicspath{{figures/}{pictures/}{images/}{./}} 
\usepackage{authblk}
\usepackage{times}                     

\usepackage{tabu}                      
\usepackage{booktabs}                  
\usepackage{lipsum}                    
\usepackage{mwe}                       

\usepackage{mathptmx}                  

\onlineid{9578}

\vgtccategory{Research}
\vgtcpapertype{algorithm/technique}
\vgtcinsertpkg

\title{R$^2$Human: Real-Time 3D Human Appearance Rendering \\from a Single Image}
\usepackage{caption}
\usepackage{graphicx}
\usepackage{multirow}
\usepackage{amsfonts}
\usepackage{bbding}
\def\etal{\emph{et al}.}
\def\eg{\emph{e.g}.}

\newcommand{\yl}[1]{{\color{black}#1}}
\newcommand{\yyw}[1]{{\color{black}#1}}
\newcommand{\yywnew}[1]{{\color{black}#1}}
\newcommand{\ylnew}[1]{{\color{black}#1}}

\author{
    Yuanwang Yang$^{1,}$\setcounter{footnote}{1}\thanks{Equal contribution.}\ ,
    Qiao Feng$^{1, \dagger}$, 
    Yu-Kun Lai$^{2}$, 
    Kun Li$^{1,}$\setcounter{footnote}{0}\thanks{Corresponding author.} \\
    $^{1}$Tianjin University, China \enspace  $^{2}$Cardiff University, United Kingdom \\
}

\teaser{
  \centering
  \includegraphics[width=\linewidth]{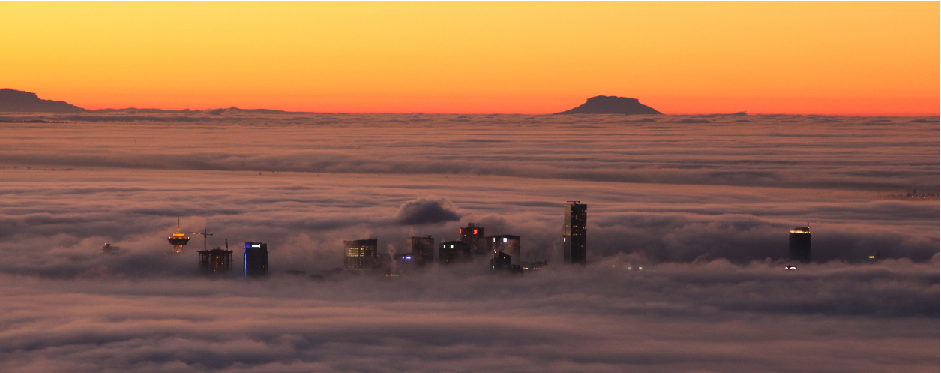}
  \caption{In the Clouds: Vancouver from Cypress Mountain.}
  \label{fig:teaser}
}

\abstract{
   \yyw{Rendering 3D human appearance from a single image in real-time} is crucial for achieving holographic communication and immersive VR/AR. Existing methods either rely on multi-camera setups or \yyw{are constrained to offline operations.}
    In this paper, we propose R$^2$Human, the first approach for real-time inference and rendering of photorealistic 3D human appearance from a single image. The core of our approach is to combine the strengths of implicit texture fields and explicit neural rendering with our novel representation, namely Z-map. Based on this, we present an end-to-end network that performs high-fidelity color reconstruction of visible areas and provides reliable color inference for occluded regions. To further enhance the 3D perception ability of our network, we leverage the Fourier occupancy field as a prior for generating the texture field and providing a sampling surface in the rendering stage. We also propose a consistency loss and a spatial fusion strategy to ensure the multi-view coherence. Experimental results show that our method outperforms the state-of-the-art methods on both synthetic data and challenging real-world images, in real-time. The project page can be found at \url{http://cic.tju.edu.cn/faculty/likun/projects/R2Human}.

}

\keywords{3D human appearance, rendering, single image, real-time}

\teaser{
  \centering
  \includegraphics[width=1.0\textwidth]{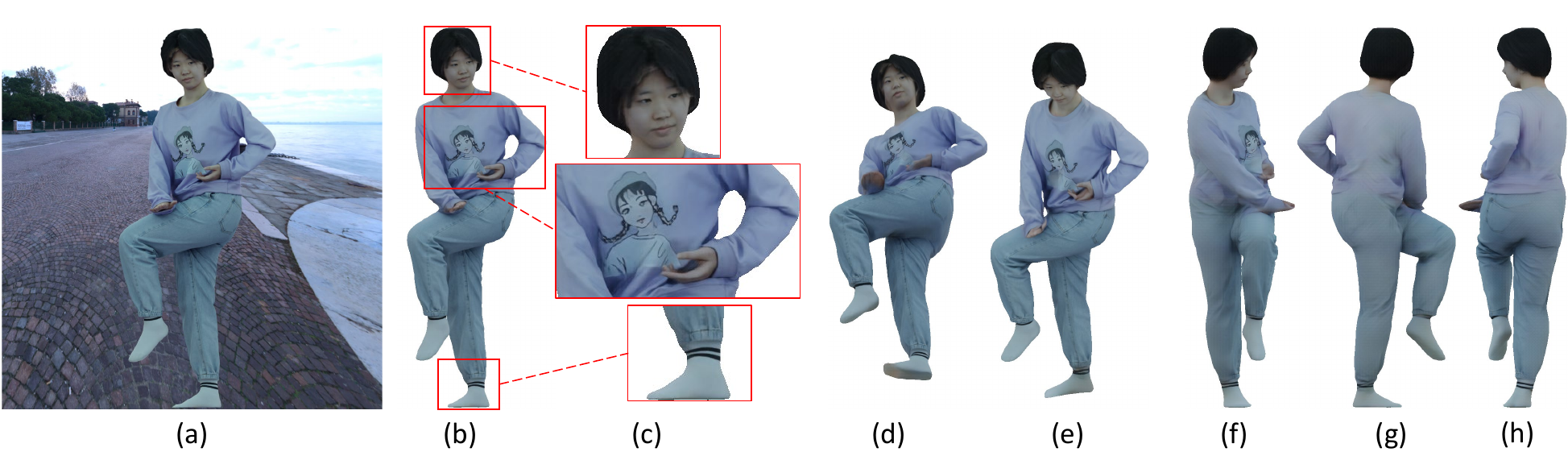}
  \caption{%
  	Given a single RGB image, R$^2$Human can generate photorealistic 3D human appearance in real-time. We utilized the input (a) to render close views (b) and zoomed in (c). The results of pitch angle changes are shown in (d-e), and (f-h) demonstrate the outcomes of divergent views.%
  }
    \label{fig:head}
}

\begin{document}


\firstsection{Introduction}

\maketitle

\yyw{Rendering 3D human appearance from a single image in real-time not only enhances visual experiences but also paves the way for practical applications in AR/VR, enabling users to engage in immersive experiences with timely feedback.} However, existing methods heavily rely on multi-camera setups~\cite{zhou2023hdhuman,shao2022floren,peng2021neural,shao2022doublefield} or are constrained to offline operations~\cite{alldieck2022photorealistic,hu2023sherf,saito2019pifu,albahar2023single}, leaving real-time 3D human rendering from a single image as an unresolved challenge. In this paper, we aim to \yyw{translate a 2D human image into a vibrant 3D appearance in real-time, facilitating holographic communication, and enhancing user experience and immersion in VR/AR.}

Research in human novel view synthesis has explored various methodologies, which can be roughly classified into two categories. The first category of methods~\cite{avidan1997novel,shao2022floren,zhou2023hdhuman} is founded on flow-based neural rendering, which warps the feature map from the input images to the target view and then produces high-quality novel view results with convolutional neural networks ({CNNs}). However, the network's predictions for occluded regions in the input image usually exhibit significant randomness due to the lack of 3D structure understanding, making it challenging to obtain consistently accurate results. To deal with this, these methods require multi-view input images to cover the rendering area as comprehensively as possible. Consequently, generating photorealistic renderings from only one image becomes difficult for them.

Another category of methods aims to recover a 3D consistent appearance for the human body. PIFu~\cite{saito2019pifu} recovers the 3D human geometry and appearance from a single human image using an occupancy field and a texture field, respectively. However, the rendered images are of low quality and the process is time-consuming. Its follow-up work~\cite{alldieck2022photorealistic} disentangles lighting from texture and redesigns the loss functions, resulting in improved visual quality. 
On the other hand, works utilizing the Neural Radiance Fields (NeRF)~\cite{mildenhall2021nerf,peng2021neural,shao2022doublefield,hu2023sherf} predict the density and radiance throughout the 3D space for rendering images by integrating along camera rays. However, NeRF-based methods often struggle with single-view inputs.
While approaches above vary in human appearance \yl{representations}, they share a common paradigm: they estimate geometry and color for discrete sampled points independently. This makes it difficult for the models to discern relationships between the sampled points, 
resulting in relatively lower quality in the synthesized novel view images. Additionally, these methods require dense spatial sampling, which results in high computational costs and makes it challenging to balance real-time performance with rendering quality.

In this paper, we introduce \textit{R$^2$Human}, a real-time framework for rendering human appearance from a single image, which uniquely combines flow-based rendering techniques with \yl{an} implicit 3D human geometry representation to synthesize novel view images. Our method only requires a single image as input and achieves both high-quality rendering performance, \yyw{as shown in Fig.\ref{fig:head}.} \yyw{Unlike previous human avatar animation typically involves creating actions beyond real-world scenarios, our focus lies in enhancing existing visual experiences by transforming real 2D human images into immersive 3D appearances in real-time. By converting existing 2D images to 3D, we cleverly avoids the increase in computational consumption caused by animation rendering and the decrease in the authenticity of the results.} Moreover, our method can generate novel view renderings of humans in various clothing without individual-specific training. We introduce an intermediate representation, \yl{namely} the Z-map, during the rendering process, which collects the source view depth of the rendered points of the target view and forms them to a 2D map. It helps \yl{lift} the 2D image feature into a 3D texture feature field, while 
\yl{maintaining}
compatibility with 2D neural rendering networks. This unique capability enables our network to learn data-driven 2D appearance knowledge of clothed human images, resulting in more accurate renderings. By preserving the features of occluded points and leveraging the Z-map, our method effectively resolves depth ambiguities.
Additionally, we employ an efficient 3D object representation known as  Fourier occupancy fields (FOF)~\cite{feng2022fof}, which explicitly represents a 3D object as a multi-channel image. It can serve both as a prior for texture field generation and as a sampling surface during the rendering stage, avoiding the high computational costs caused by dense sampling. 
\yyw{In order to ensure the consistency between multiple views and reduce the jittering phenomenon between neighboring views, we propose a consistency loss to 
\ylnew{regularize}
this process and design a spatial fusion strategy to enhance this in practical applications.}
Our method paves the way for the practical \yl{applications} of AR/VR, which can be applied in holographic communication in the future. \emph{Source code will be available for research purposes}.

In summary, the contributions of our work are as follows:
\begin{itemize}
    \item We present a novel system for high-quality, real-time synthesis of human novel view images with only a single RGB image input. To the best of our knowledge, this is the first system to restore the full-body appearance of a 3D human in \ylnew{real-time} from a single image, \yyw{paving the way for practical applications in AR/VR.}
    \item We propose \textit{R$^2$Human}, an end-to-end CNN-based neural rendering method that combines the strengths of implicit texture field and explicit neural rendering, which can produce results with both 3D consistency and high visual quality.
    \item We introduce an intermediate representation called Z-map, which alleviates depth ambiguities in rendering, enabling high-fidelity color reconstruction for the visible area while providing reliable color inference for the occluded regions.
    \yyw{\item We propose a consistency loss to ensure the multi-view coherence and reduce the jittering phenomenon, and design a spatial fusion strategy to enhance this in practical applications.}
\end{itemize}

\section{Related work}

\begin{table}[!t]
\centering
\caption{Comparison with state-of-the-art methods. \XSolidBrush: not supported, \Checkmark: supported.}
\label{comparison}
\resizebox{\linewidth}{!}{%
\begin{tabular}{ccccc}
\toprule[1.5pt]
\multirow{2}{*}{Method} &Monocular &Real-time &Fully-body &High-quality\\
&Input &Processing &Rendering &Output \\
\specialrule{0.1em}{1pt}{2pt}
Project Starline~\cite{lawrence2021project} &\XSolidBrush &\Checkmark &\XSolidBrush &\Checkmark\\
3DTexture~\cite{waechter2014let} &\XSolidBrush &\XSolidBrush &\Checkmark &\XSolidBrush\\
Floren~\cite{shao2022floren} &\XSolidBrush &\XSolidBrush &\Checkmark &\Checkmark\\
HDHuman~\cite{zhou2023hdhuman} &\XSolidBrush &\XSolidBrush &\Checkmark &\Checkmark\\
PIFu~\cite{saito2019pifu} &\Checkmark &\XSolidBrush &\Checkmark &\XSolidBrush\\
SHERF~\cite{hu2023sherf} &\Checkmark &\XSolidBrush &\Checkmark &\XSolidBrush\\
Ours &\Checkmark &\Checkmark &\Checkmark &\Checkmark\\
\bottomrule[1.5pt]
\end{tabular}%
}
\end{table}

\subsection{Monocular 3D Human Reconstruction}
Human reconstruction has long been a concern in the domain of computer vision and graphics. Methods for predicting human bodies from a single image represent 3D shapes primarily by estimating the properties of points in 3D space. Some previous works~\cite{zheng2019deephuman,varol2018bodynet} directly predict the occupancy field of a given segment in space through regression networks, but such methods have high memory requirements, which limits the spatial resolution of shape estimation. Another type of methods involves using implicit function networks to remove resolution constraints. Saito \etal~\cite{saito2019pifu} proposed an implicit function based on pixel-aligned features to reconstruct a 3D human from a single image. PIFuHD~\cite{saito2020pifuhd} further introduces normal maps to improve geometric details. Subsequent methods~\cite{zheng2021pamir,xiu2022icon,xiu2022econ,zhang2023joint2human} improve the robustness of the results by incorporating parametric SMPL~\cite{smpl} priors. \yyw{Albahar \etal~\cite{albahar2023single} combine diffusion model to obtain detailed geometry and texture, but the reconstruction effect of loose clothing is not good. Some methods~\cite{zielonka2023drivable,guo2023vid2avatar,jiang2022selfrecon,weng2022humannerf,jiang2022neuman,jiang2023instantavatar} take advantage of the features of SMPL to align different poses to obtain detailed reconstructions from video, but cannot adapt to single image settings.} Although these methods can achieve realistic results, they often require significant computational resources and are time-consuming. Feng \etal~\cite{feng2022fof,feng2022monocular} proposed an efficient 3D geometric representation called Fourier Occupancy Field, which can establish a strong link between 3D geometry and 2D images, significantly reducing the computational requirements for reconstruction. However, one limitation is \yl{its} inability to estimate texture during the geometry reconstruction process. \yyw{Other work~\cite{song2023rc,lu20233d} relies on RGBD cameras to provide depth information, thus ensuring real-time performance and reconstruction accuracy. But their results rely too much on the SMPL model, making it impossible to adapt to a variety of clothes, and the use of RGBD cameras also limits its application in daily life.}

\begin{figure*}[htb]
    \centering
    \includegraphics[width=.95\textwidth]{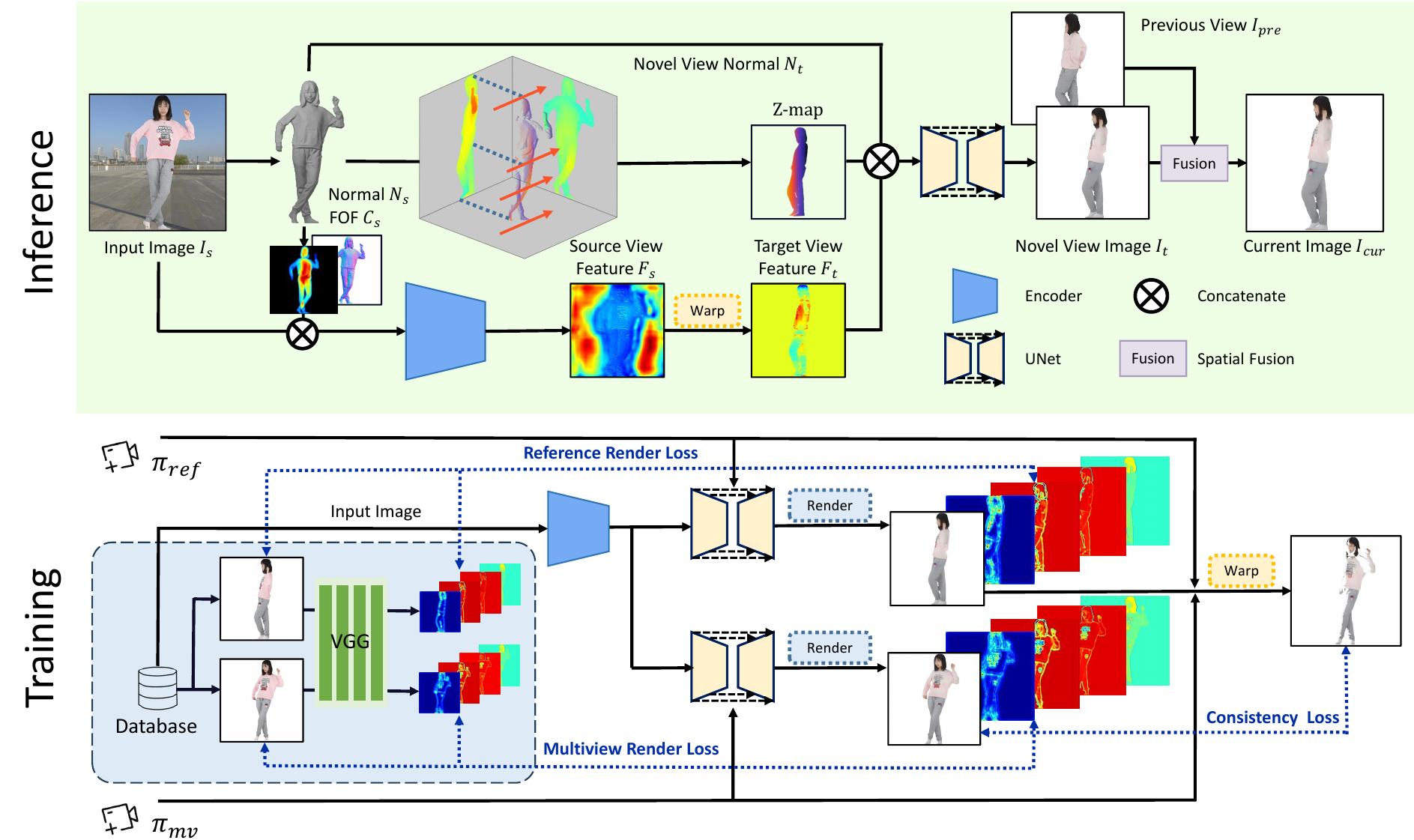}
    \captionsetup{type=figure}
    \captionof{figure}{
    \yyw{
    The overall pipeline of \textit{R$^2$Human} for real-time 3D human appearance rendering. \textit{R$^2$Human} leverages the proposed Z-map to combine the strengths of implicit texture field and explicit neural rendering seamlessly. With our consistency loss, we constrain the rendered color of the same visible point to be consistent across different views, thereby ensuring the multi-view consistency of the results.}
}
    \label{fig:main}
\end{figure*}

\subsection{Human Novel View Synthesis}
\yl{Flow}-based neural rendering \yl{methods} can synthesize realistic novel human view images. Shao \etal~\cite{shao2022floren} estimate robust appearance flow with epipolar constraints, reduce ambiguous texture warping, and then synthesize high-quality novel human perspective. Zhou \etal~\cite{zhou2023hdhuman} first estimate highly detailed 3D human geometry, and then effectively solve the serious occlusion problem caused by sparse views through geometry-guided pixel-wise feature integration method. Although these methods reduce the impact of occlusion on flow-based texture warping, they still require multi-view information to ensure the robustness of the results. Phong \etal~\cite{nguyen2022free} warp images from a single sparse RGB-D input by sphere-based rendering and refine the resulting image using an additional occlusion-free input, but its application \yl{is} limited by its reliance on depth sensors. On the other hand, some methods~\cite{saito2019pifu,xiu2022icon,yang2023d} that focus on geometry can predict textures \yl{during reconstruction}, and then \yl{synthesize} novel views through traditional rendering methods. Thiemo \etal~\cite{alldieck2022photorealistic} improved the visual fidelity of the results by computing albedo and shading information and carefully designing rendering losses. However, these methods find it difficult to learn the relationships between points in space due to the use of discrete sampling points to estimate color, resulting in relatively poor rendered image quality. 

In recent years, neural radiance fields (NeRF)~\cite{mildenhall2021nerf} have become a powerful novel view synthesis tool, using multi-layer perceptrons (MLP) to model a scene as density and color fields, resulting in realistic rendering. Researchers have made many attempts on the basis of NeRF. Peng \etal~\cite{peng2021neural} \yl{anchor} the learned potential encoding onto SMPL~\cite{smpl} fixed points to integrate \yl{multi-view} video information, but it can only render humans with relatively uniform textures. Shao \etal~\cite{shao2022doublefield} combine the surface field and the \yl{radiance} field to achieve high-quality rendering, but are still limited by the need for multiple views. Hu \etal~\cite{hu2023sherf} proposed the first generalizable Human NeRF based on a single image input, but visible artifacts still exist in partially occluded bodies in the observation space. In addition, these implicit representation based methods all have a common problem: due to the high computational complexity required for dense sampling, they 
\yl{are time-consuming.}

In this paper, we contribute the first real-time and high-quality human novel view synthesis system from a single RGB image, which combines the strengths of implicit texture field and explicit neural rendering \yyw{and translates a 2D human image into a vibrant 3D appearance in real-time, facilitating holographic communication, and enhancing user experience and immersion in VR/AR.}
\yyw{
Tab.~\ref{comparison} presents a comparative analysis focusing on key features: support for monocular input, real-time processing capabilities, full-body rendering, and high-fidelity novel view synthesis. As indicated in the table, our method is unique in its support for all these features, distinguishing it from other approaches.}



\section{Method}
\label{method}

Our goal is to build an end-to-end trainable framework that can achieve photorealistic novel view rendering of humans from a single RGB image. We employ FOF~\cite{feng2022fof} to represent the 3D human form as a multi-channel image, thereby reducing computational time and memory overhead. 
Additionally, we introduce Z-map to effectively address the issue of prediction ambiguity for occluded points within the input image. Consequently, our network is capable of producing high-quality renderings of novel views using just a single image.

As shown in Fig.~\ref{fig:main}, our framework mainly consists of two parts: 
1) \yl{We develop an} image encoder to generate a texture field aligned with the reconstructed 3D geometry. 
2) \yl{We warp} the texture field and obtain other priors, \yl{and} then use a rendering network to synthesize results with high visual quality.
At inference, we estimate a detailed mesh from the input image, and using a pixel-aligned feature encoder to extract texture features. Next we mitigate depth blur in rendering with the help of Z-map representations, resulting in high-quality textures. During training, we use the input image to render two supervised views. We warp the first image to the second and perform visibility inference. By supervising the color of the visible points, we ensure the multi-view consistency of the results.

\subsection{Z-map}
\label{zmap}
\subsubsection{Definition of Z-map}
\yyw{We propose a new representation called Z-map, which combines the strengths of flow-based rendering with implicit-field-based rendering. 
Unlike existing approaches that utilize the depth-map of the novel view for appearance rendering, our Z-map comprises the source view $z$-coordinate of the visible points in the novel view, aligned to the novel view.
By leveraging the source view image features and the $z$-coordinate, the Z-map uniquely determines the position of visible points in the implicit field, thus eliminating depth ambiguity. }

Most existing methods \yl{taking} a single image as input rely on \yl{a} pixel-aligned implicit function to produce a 3D field, such as a texture field. The color of a point in 3D space can be predicted by:
\begin{equation}
    \label{33}
    c = f_{mlp}(e, z),    
\end{equation}
where $c\in \mathbb{R}^3$ is the predicted RGB color, $e$ is the corresponding feature sampled from the image feature map, $z$ is the source view depth of that point, and $f_{mlp}$ is the MLP-based decoder. In the task of view synthesis, each pixel on the output image $I$ corresponds to a point in the 3D space. Thus, we can collect the colors $c$ for all those points to generate the final result.

Such a schema does not consider the internal relationship between those points. \yl{In contrast,} flow-based methods decode those features simultaneously \yl{to avoid} such a drawback, which can be written as:
\begin{equation}
    I = f_{cnn}(F_{map}(\mathcal{F}_s)),    
\end{equation}
where $\mathcal{F}_s$ is the feature map of the source view, $F_{map}$ is the wrapping flow, and $f_{cnn}$ is the CNN-based decoder. 

Let $\mathcal{F}_t=F_{map}(\mathcal{F}_s)$ be the feature map corresponding to the target view. We can notice that $\mathcal{F}_t$ is the collection of $e$ in Eq. \ref{33}. Therefore, if we collect all those source view depth values and form them into a 2D Z-map, these two kinds of methods can be unified in a single framework:
\begin{equation}
    I = f_{cnn}(\mathcal{F}_t, Z_{map}),    
\end{equation}
where $Z_{map}$ is the Z-map. Like the warping flow, Z-map can also be directly produced, which is more efficient than calculating $z$ values for each point separately.

\subsubsection{Calculation of Z-map}
To combine the strengths of implicit texture field and explicit neural rendering, we propose the Z-map, which enables rendering networks to achieve 3D perception and avoid one-to-many problems caused by depth ambiguity. We define Z-map as the depth obtained from the 
\yl{transformation}
of visible points in the novel view to the source view, 
so we can obtain Z-map while calculating the flow maps. Specifically, we render the depth maps $D_{s},D_{t}$ first for both views using the predicted mesh based on camera parameters. Then, for each coordinate point $v_{t} \in \mathbb{R}^2$ in the novel view, we inverse project it to the world coordinate system using the camera parameters, and then project it to the source view space to obtain flow maps and Z-map from the novel view to the source view:
\begin{equation}
  \label{66}
  (F_{map},Z_{map}) = \Pi^{s}((\Pi^{t})^{-1}(v_{t},D_{t})),
\end{equation}
where $\Pi^{s}$ is the projection matrix transforming points from the world \yl{coordinates} to the source view \yl{coordinates} $v_{s}$ and ($\Pi^{t})^{-1}$ is the matrix transforming points from the novel view \yl{coordinates} $v_{t}$ to the world \yl{coordinates}.

\subsection{R$^2$Human Networks}
\label{network}

\subsubsection{Pixel-\yl{aligned} Feature Encoder} To leverage the geometric information obtained from the reconstruction network and enhance feature extraction, our encoder incorporates additional information derived from the reconstructed geometry. This allows for a more comprehensive understanding of the input image. This approach consists of two key components: FOF and normal map.

\noindent\textbf{FOF.} As shown in Fig.~\ref{fig:main}, we obtain FOF through the existing method. It is a multi-channel image that contains \yl{3D} information and enables the encoder to capture and represent spatial relationships, depth cues, and other geometric aspects inherent in the Human. This can lead to a more robust and information-rich representation of features, potentially improving the performance of subsequent rendering.

\noindent\textbf{Normal map.} We generate a normal map by calculating the surface normals for each point in the predicted mesh. It can \yl{enhance} the encoder's ability to perceive lighting information and geometric details.

In summary, our encoder network E is defined as follows:
\begin{equation}
  \mathcal{F}_{s} = E(\oplus(I_{s},C_{s},N_{s})),
\end{equation}
where $\oplus$ is the concatenation operation. $I_{s},C_{s},N_{s}$ are the input color image, the predicted FOF and estimated surface normal map, respectively. 

\subsubsection{Novel View Rendering} Similar to \yl{the} encoder, our rendering network also integrates some additional information to improve rendering performance. Note that we do not integrate Fourier occupation fields because rendering only focuses on the visible part of the image. We use novel camera parameters $\Pi^{t}$ to render the normal map. 
Then we warp the features of the source view to the novel view based on the flow map:
\begin{equation}
  \mathcal{F}_{t}(v) = \mathcal{F}_{s}(F_{map}(v)),
\end{equation}
In summary, our \yl{rendering} network $R$ is defined as follows:
\begin{equation}
    I_t = R(\oplus(\mathcal{F}_t,Z_{map},N_t)),
\end{equation}
where $N_{t}$ is the normal map rendered by novel camera $\Pi^{t}$.

\yyw{
\subsubsection{Spatial Fusion Strategy}
To maintain consistency during the rotation of the view, we introduce a spatial fusion strategy for generating the free-view video stream. Specifically, during the rotation process, we keep the image of the previous view $I_{pre}$, and use Eq. \ref{77} 
to get the surface points $v$ that are visible in both the current and previous views. Subsequently, the color of $v$ in the current view image $I_{cur}$ is interpolated as:}
\begin{equation}
    I_{cur}(v) = \alpha I_{t}(v) + (1-\alpha) I_{pre}(v),
\end{equation}
\yyw{where $\alpha=\beta |sin(\phi/2)|$, $\phi$ is the rotation angle of the current view relative to the input view. 
The closer the current view is to the input view, the more we prefer the color of $I_{t}$, 
while the farther the current view is from the input view, the more $I_{pre}$ we use to ensure consistency. We set $\beta$ to ${0.8}$ in our experiments.}

\yyw{\subsection{Training}}
\yyw{We use synthetic data to train R$^2$Human during the training stage. As illustrated in Fig. \ref{fig:main}, an input image along with its corresponding FOF and normal map, sampled from the training set, serves as the input. The encoder then generates the corresponding texture field features, denoted as $\mathcal{F}$. With the camera parameters $\Pi$ (comprising camera direction and position), we compute the 
\ylnew{warping}
flow which contains the novel view information we need, enabling the generation of an image of any view with the decoder. 
In each gradient step, we synthesize two images $\hat{I}_{ref}$, $\hat{I}_{mv}$ with the same texture field features from a reference camera $\Pi^{ref}$ and an additional camera $\Pi^{mv}$ to perform multi-view consistency supervision. 

\vspace{+0.1cm}
\noindent\textbf{Consistency Loss.} To guarantee consistency between multiple views and reduce the jittering phenomenon during the rotation of the view, we propose a consistency loss to supervise the multi-view images rendered from the same texture field features. Specifically, we use Eq. \ref{66} to calculate the flow map between the reference camera $\Pi_{ref}$ and the multi-view camera $\Pi_{mv}$. Then we warp the image $\hat{I}_{ref}$ rendered by the reference camera to the muti-view camera view to get $\hat{I}^{mv}_{ref}$. Visibility inference is then performed: If the difference between the projection depth of reference view $D^{mv}_{ref}=\Pi^{mv}((\Pi^{ref})^{-1}(v,D_{ref}))$ and the multi-view depth $D_{mv}$ is lower than a threshold, we treat coordinate point v as visible: 
\begin{equation}
    \label{77}
    |D^{mv}_{ref}(v) - D_{mv}(v)| < \lambda min(D^{mv}_{ref}(v), D_{mv}(v)),
\end{equation}
where $\lambda$ is a hyper-parameter and we set it to 0.02 in all our experiments. 

We use the $L_1$ loss to supervise the visible points in the reference camera and the multi-view camera. The loss function $L_{consistency}$ can be expressed as:
\begin{equation}
    L_{consistency}=\frac{1}{|M^{mv}_{ref}|}\sum_{(x,y) \in M^{mv}_{ref}} \Vert \hat{I}^{mv}_{ref}(x,y)-\hat{I}_{mv}(x,y)\Vert_1,
\end{equation}
where $M^{mv}_{ref}$ is the set of visible points between reference camera and multi-view camera calculated by Eq. \ref{77}.

\vspace{+0.1cm}
\noindent\textbf{Pixel Loss.} Let the ground truth of the reference view and multi-view be $I_{ref}$ and $I_{mv}$. We add \ylnew{an} $L_1$ \ylnew{loss} to supervise the pixel points between both sets of pairs (${I}_{ref}$, $\hat{I}_{ref}$) and (${I}_{mv}$, $\hat{I}_{mv}$), and to make the network more focused on the human, we only supervise the human foreground region of the image:
\begin{equation}
    L_{pixel}=\frac{1}{|M|}\sum_{(x,y) \in M} \Vert I(x,y)-\hat{I}(x,y)\Vert_1,
\end{equation}
where $M$ is the set of foreground pixels of $I$.

\vspace{+0.1cm}
\noindent\textbf{LPIPS Loss.} In addition, to enhance the visual effect of the output image, we also add the LPIPS loss described in ~\cite{zhang2018unreasonable} for both sets of pairs (${I}_{ref}$, $\hat{I}_{ref}$) and (${I}_{mv}$, $\hat{I}_{mv}$):
\begin{equation}
    L_{LPIPS}=\frac{1}{|M|}\sum_{(x,y) \in M} \Vert VGG(I(x,y))-VGG(\hat{I}(x,y))\Vert_2,
\end{equation}
where $VGG$ is the network described in ~\cite{simonyan2014very}.

Finally, our loss can be formulated as follows:
\begin{equation}
    L = \lambda_{1} L_{consistency} + \lambda_{2} L_{pixel} + \lambda_{3} L_{LPIPS},
\end{equation}
where we set $\lambda_{1}$, $\lambda_{2}$ and $\lambda_{3}$ to 100, 1 and 0.5 in our experiments. 
}

\vspace{+0.1cm}
\noindent\textbf{Implement Details.} Our our \yl{network} is trained on synthetic data \yl{including} pairs of meshes and rendered images. We collect 526 high-quality human scans from THuman2.0 dataset~\cite{Thuman} with a wide range of clothing, poses and shapes. We randomly split them into a training set of 368 scans and a testing set of 105 scans. The remaining subjects are used as the validation set. 
\yl{For training our network, instead of using FOF constructed from the ground truth geometry, we 
apply the FOF-SMPL reconstruction network~\cite{feng2022fof} to predict the FOF representation from the single-view input image.}
Using the predicted FOF for training can make the rendering network obtain stronger generalization ability and \yl{improved} robustness. 
We implement our method using PyTorch and train all 
network components jointly, end-to-end, using the Adam optimizer, with \yl{a} learning-rate of $2 \times 10^{-4}$. We train the \yl{network} for 10 epochs which takes about 4 days with a batch size of 4 using a single NVidia RTX3090 GPU.

\section{Experiments}

\subsection{Comparisons}
\noindent\textbf{Metrics.} We use three widely used metrics for images to quantitatively evaluate our method: structural similarity (SSIM)~\cite{wang2004image}, peak signal-to-noise ratio (PSNR)~\cite{sara2019image}, and learned perceptual image patch similarity (LPIPS)~\cite{zhang2018unreasonable} that uses AlexNet~\cite{krizhevsky2012imagenet} to extract features.


\vspace{+0.1cm}
\noindent\textbf{Baselines.} We quantitatively compare our approach with two methods~\cite{saito2019pifu,hu2023sherf}: 1) PIFu~\cite{saito2019pifu} uses implicit functions based on pixel-aligned features to reconstruct 3D humans from a single image, and we use 
\yl{a standard rendering pipeline}
to render it to obtain a \yl{novel} view image. 2) SHERF~\cite{hu2023sherf} is the first generalizable Human NeRF model to recover 3D humans from a single image, achieving state-of-the-art performance compared with previous generalizable Human NeRF methods. Additionally, to showcase the necessity of utilizing neural rendering, we conducted qualitative comparisons with state-of-the-art texture mapping methods 3DTexture~\cite{waechter2014let}.

\begin{figure*}
    \centering
    \includegraphics[width=.73\textwidth]{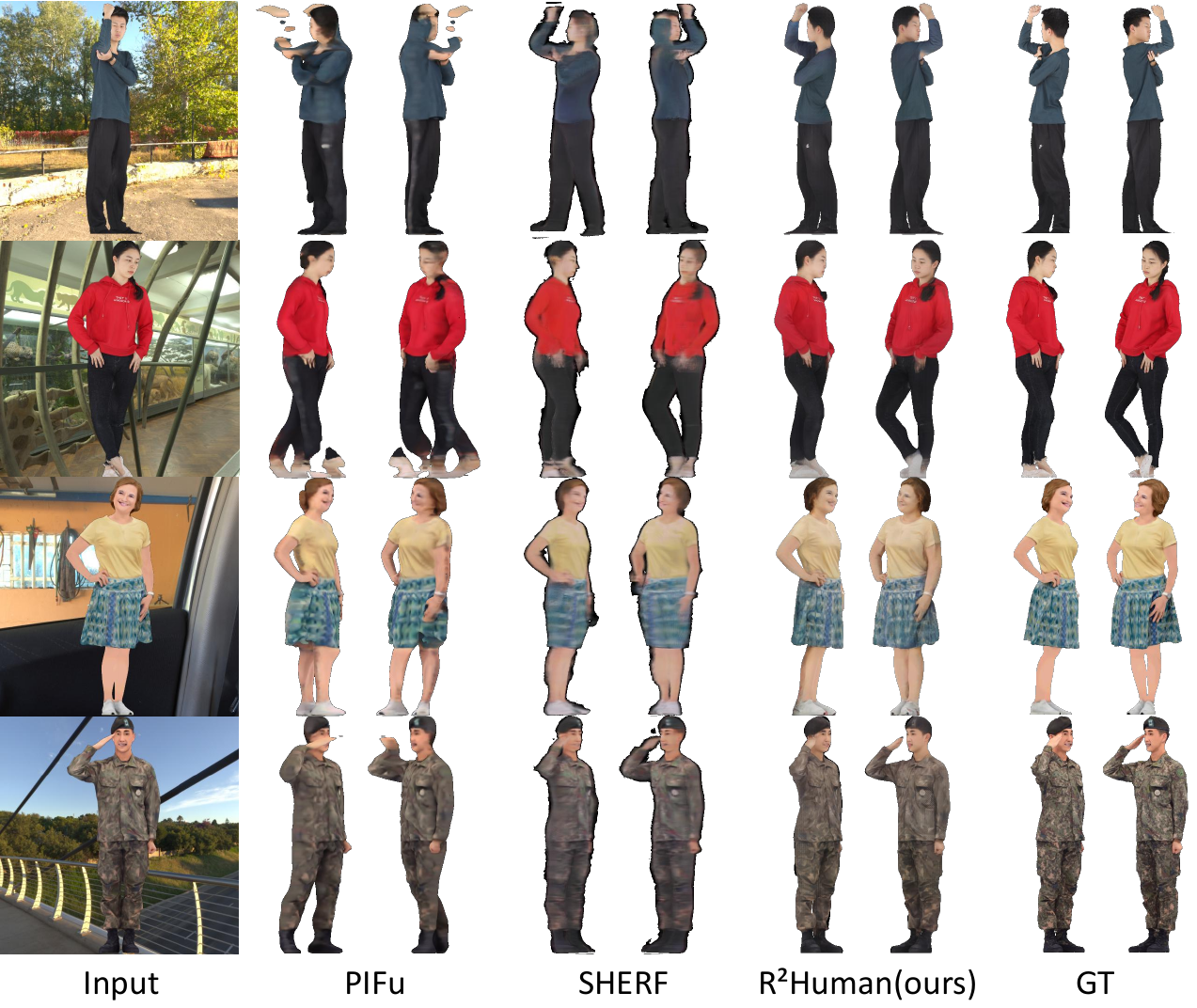}
    \captionsetup{type=figure}
    \captionof{figure}{Novel view rendering on THuman2.0 (top tow rows) and 2k2k dataset (bottom tow rows). 
    }
    \label{fig:qualitative_thuman}
\end{figure*}

\begin{figure}[htb]
    \centering
    \includegraphics[width=.4\textwidth]{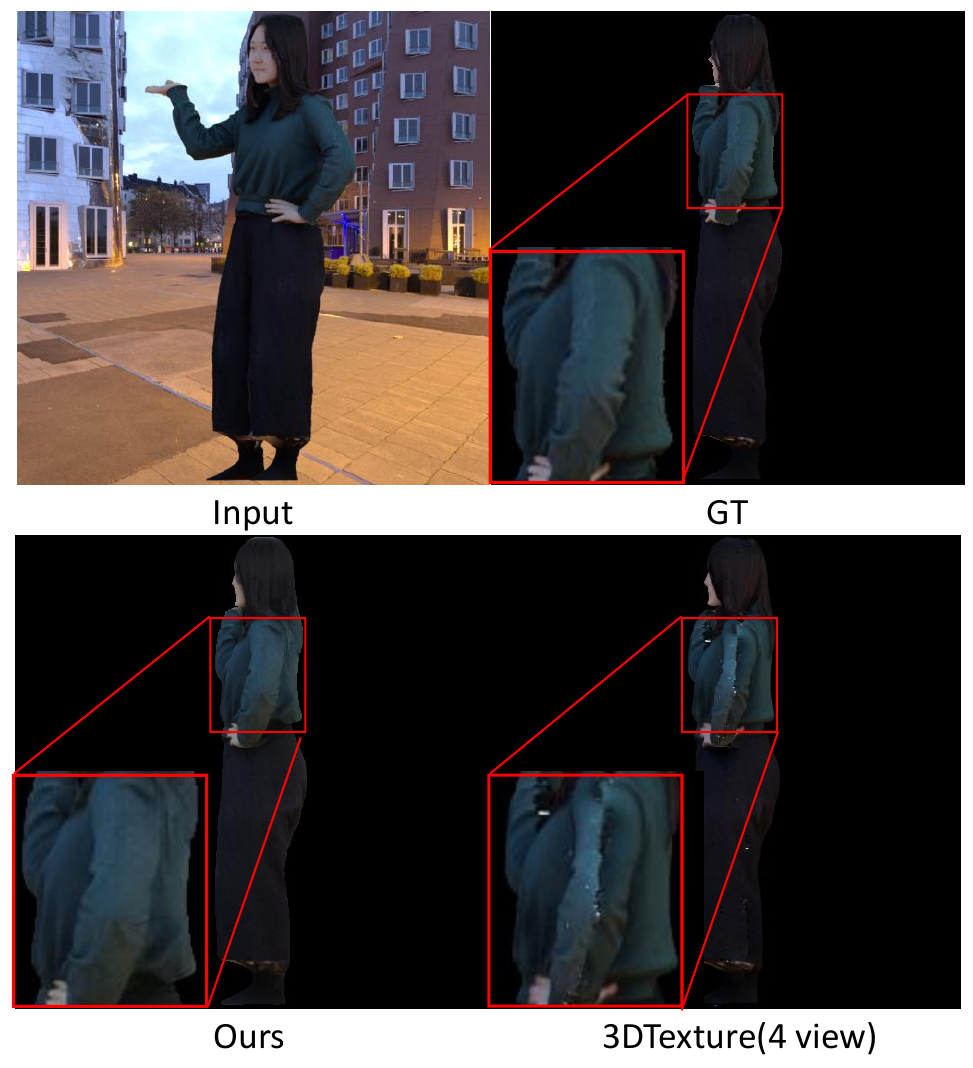}
    \captionsetup{type=figure}
    \captionof{figure}{Qualitative comparison with 3DTexture. The results show that even with four views, our single-view approach still outperforms traditional texture mapping.}
    \label{fig:noflow}
    \vspace{-0.5cm}
\end{figure}

\vspace{+0.1cm}
\noindent\textbf{Qualitative results.} Fig.~\ref{fig:qualitative_thuman} and Fig.~\ref{fig:qualitative_2k2k} presents qualitative comparisons of our approach to the baseline \yyw{on the 2k2k dataset~\cite{han2023high} and the THuman2.0 dataset~\cite{Thuman}, respectively.} PIFu demonstrates the ability to estimate reasonably accurate colors, but its performance is hindered by limitations in geometric estimation, resulting in suboptimal image quality. 
\yyw{On the other hand, while SHERF can produce results with accurate pose, it encounters challenges in generating desirable outcomes for individuals wearing loose clothing, mainly due to its heavy reliance on SMPL priors. Additionally, since SHERF estimates colors for sampling points independently, it does not effectively consider the interrelationship between these points, often leading to ambiguous results.}
Fig.~\ref{fig:noflow} shows a qualitative comparison with traditional texture mapping methods. 3DTexture has a noticeable concatenation \yl{artifacts} in the multi-view settings and cannot render occluded areas in the input view. Thanks to Z-map, R$^2$human is able to incorporate geometric priors to enhance the rendering of occluded areas by 3D information. Furthermore, with the aid of neural rendering, R$^2$human can effectively compensate for any inaccuracies in the geometric estimates. As a result, R$^2$human is capable of synthesizing high-fidelity results that exhibit remarkable 3D consistency. 

\vspace{+0.1cm}
\noindent\textbf{Quantitative evaluations.} We selected 105 models in the \yyw{2k2k} and THuman2.0 test sets for evaluation, respectively. Tab.~\ref{quantitative_little} shows the performance of the method when the novel view is in close proximity to the source view (with a difference of $\leq$ 45°), and Tab.~\ref{quantitative_big} shows the performance when the novel view is significantly different from the source view (with a difference of $\geq$ 90°). It is evident that R$^2$Human outperforms SHERF and PIFu across all evaluation metrics, regardless of whether there are minor or major changes in the viewing angle.

\begin{table}[h]
    \centering
    \caption{Quantitative rendering results in close views.}
    \label{quantitative_little}
    \resizebox{.5\textwidth}{!}{%
    \begin{tabular}{ccccccc}
        \toprule[1.5pt]
        \multirow{2}{*}{Method} &\multicolumn{3}{c}{THuman2.0} &\multicolumn{3}{c}{2k2k}\\ 
        \cmidrule(lr){2-4}\cmidrule(lr){5-7}
                                  &SSIM$\uparrow$   &PSNR$\uparrow$  &LPIPS$\downarrow$ &SSIM$\uparrow$   &PSNR$\uparrow$  &LPIPS$\downarrow$\\ 
        \hline
        \multirow{1}{*}{PIFu~\cite{saito2019pifu}}    &0.8576 &19.45 &0.1413                 &0.8648 &20.40 &0.1139      \\
        
        \multirow{1}{*}{SHERF~\cite{hu2023sherf}}     &0.8972 &23.15 &0.1078                 &0.8864 &22.23 &0.1113        \\
        
        \multirow{1}{*}{Ours}      &\textbf{0.9415} &\textbf{26.92} &\textbf{0.0478} &\textbf{0.9134} &\textbf{24.69} &\textbf{0.0656}\\
        \bottomrule[1.5pt]
    \end{tabular}%
}
\end{table}
\begin{table}[h]
    \centering
    \caption{Quantitative rendering results in divergent views.}
    \label{quantitative_big}
    \resizebox{.5\textwidth}{!}{%
    \begin{tabular}{ccccccc}
        \toprule[1.5pt]
        \multirow{2}{*}{Method} &\multicolumn{3}{c}{THuman2.0} &\multicolumn{3}{c}{2k2k}\\ 
        \cmidrule(lr){2-4}\cmidrule(lr){5-7}
                                  &SSIM$\uparrow$   &PSNR$\uparrow$  &LPIPS$\downarrow$ &SSIM$\uparrow$   &PSNR$\uparrow$  &LPIPS$\downarrow$\\ 
        \hline
        \multirow{1}{*}{PIFu~\cite{saito2019pifu}}      &0.8409 &18.68 &0.1681       &0.8441 &19.14 &0.1447      \\
        
        \multirow{1}{*}{SHERF~\cite{hu2023sherf}}     &0.8808 &21.68 &0.1249         &0.8708 &20.91 &0.1283        \\
        
        \multirow{1}{*}{Ours}      &\textbf{0.9221} &\textbf{25.36} &\textbf{0.0660} &\textbf{0.8889} &\textbf{22.75} &\textbf{0.0914}\\
        \bottomrule[1.5pt]
    \end{tabular}%
}
\end{table}

\yyw{\noindent\textbf{Comparison of running times.} Tab. \ref{tab:times} shows the comparison results in terms of running time. We use TensorRT to accelerate inference in the real-time system. It can be seen that our method runs significantly faster than the other two baseline methods, with a speed improvement of two orders of magnitude.}

\begin{table}[ht]
    \centering
    \caption{Comparison of Running Times.}
    \label{tab:times}
    \resizebox{.95\linewidth}{!}{%
    \begin{tabular}{cccccc}
    \toprule[1.5pt]
     &PIFu &SHERF &Ours &Ours (TensorRT)
    \\
    \specialrule{0.1em}{1pt}{2pt}
    Time (ms) &5741.62 &1257.90 &92.09 &\textbf{13.13}
    \\
    \bottomrule[1.5pt]
    \end{tabular}%
    }
\end{table}

\begin{figure}[htb]
    \centering
    \includegraphics[width=0.4\textwidth]{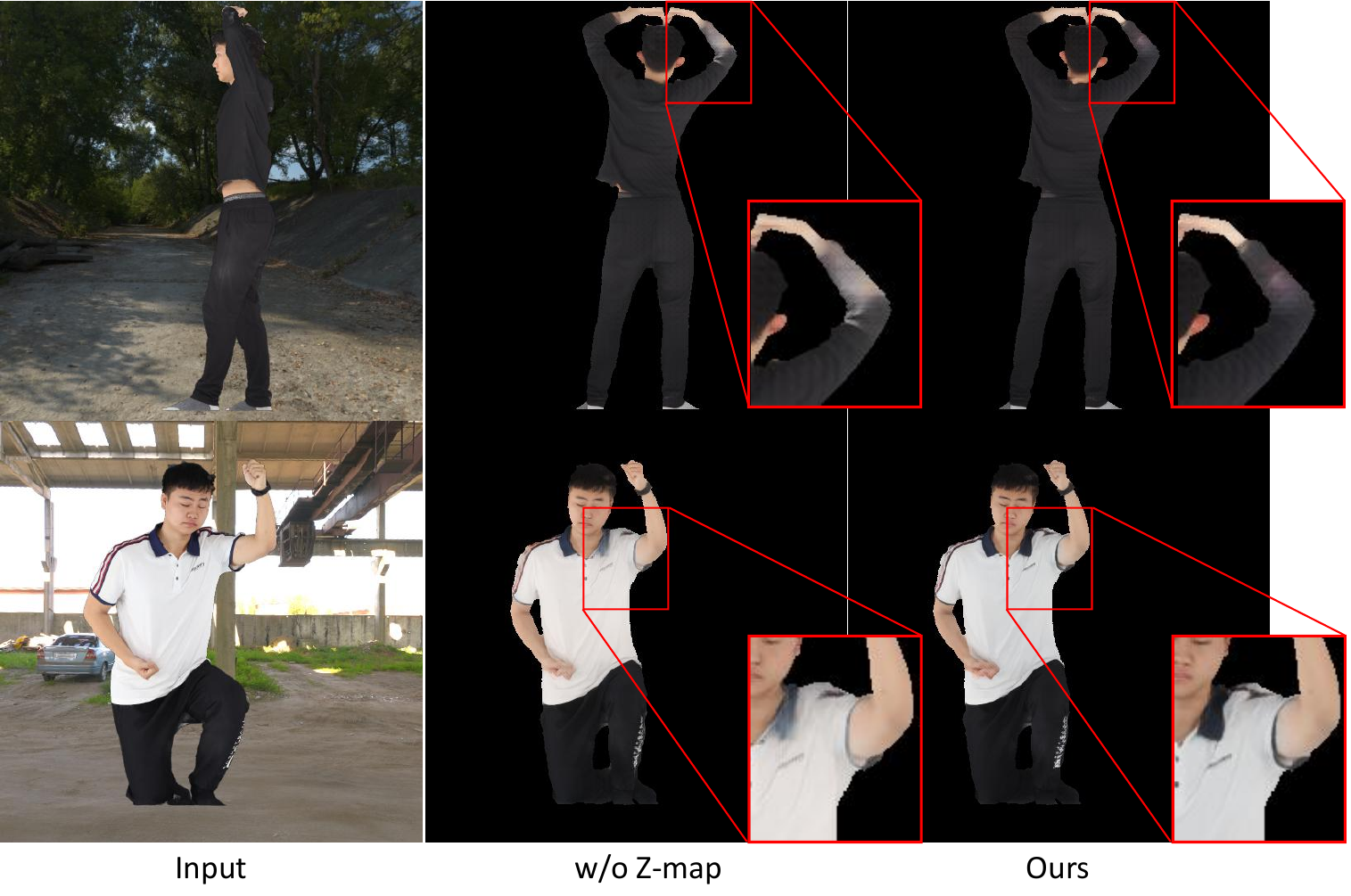}
    \captionsetup{type=figure}
    \captionof{figure}{Ablation study comparing our model with and \yl{without} the Z-map in the decoder.}
    \label{fig:ablation_zmap}
\end{figure}

\begin{figure}[htb]
    \centering
    \includegraphics[width=0.47\textwidth]{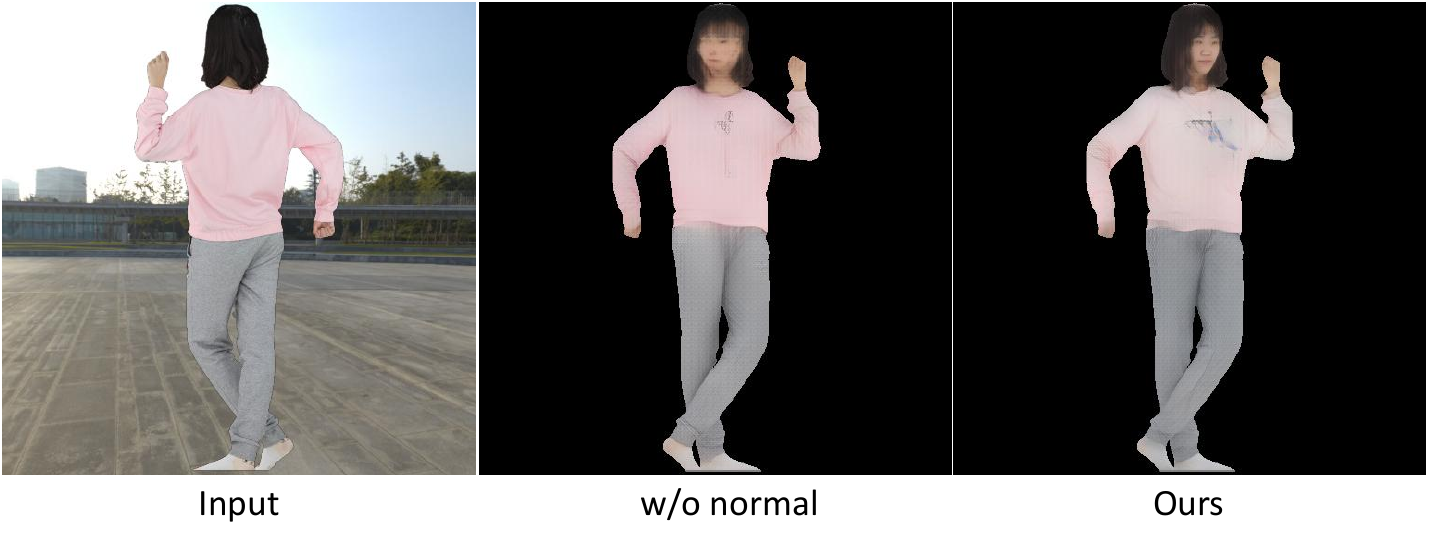}
    \captionsetup{type=figure}
    \captionof{figure}{Ablation study comparing our model with and \yl{without} the normal in the \yyw{rendering network}.}
    \label{fig:ablation_normal}
    \vspace{-0.5cm}
\end{figure}

\begin{figure}[htb]
    \centering
    \includegraphics[width=0.4\textwidth]{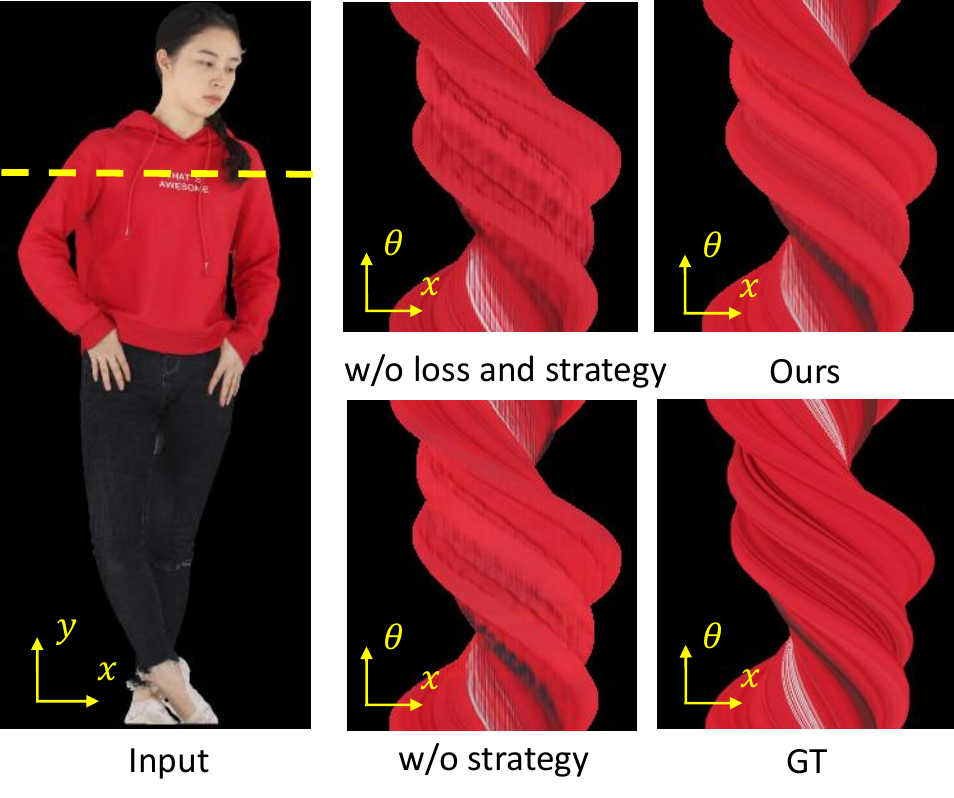}
    \captionsetup{type=figure}
    \captionof{figure}{\yyw{Ablation study on consistency loss in training and spatial fusion strategy in inference.}}
    \label{fig:ablation_epi}
\end{figure}

\begin{figure}[htb]
    \centering
    \includegraphics[width=0.5\textwidth]{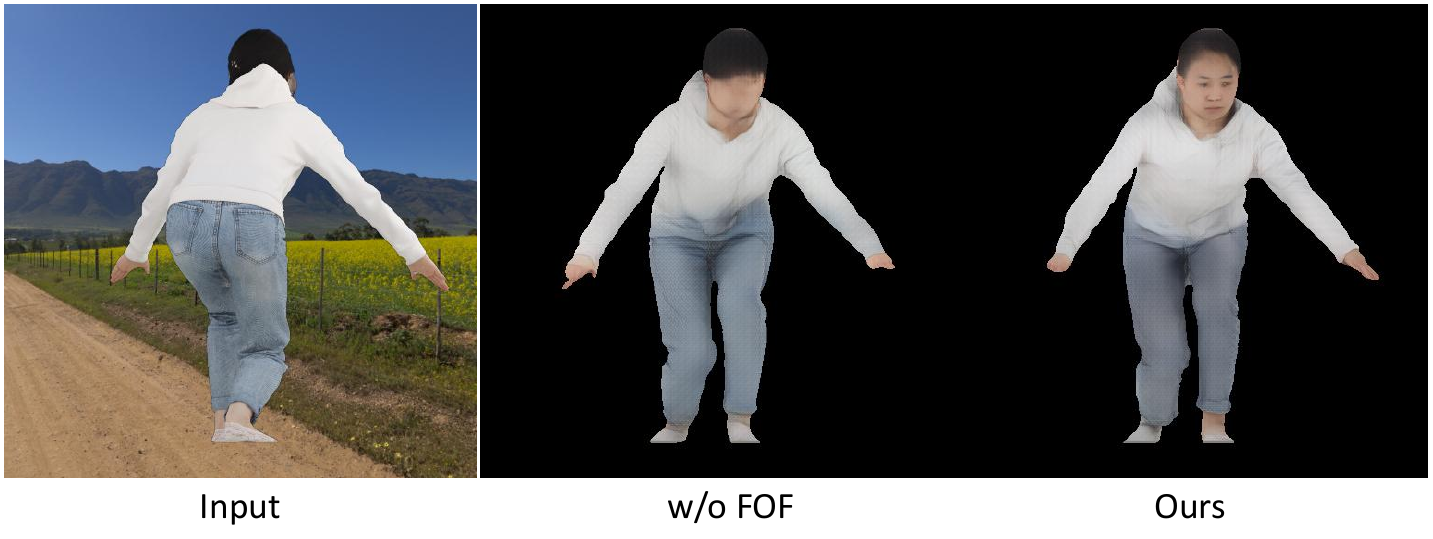}
    \captionsetup{type=figure}
    \captionof{figure}{Ablation study comparing our model with and \yl{without} the FOF in the encoder.}
    \label{fig:ablation_fof}
\end{figure}

\begin{figure}[htb]
    \centering
    \includegraphics[width=0.5\textwidth]{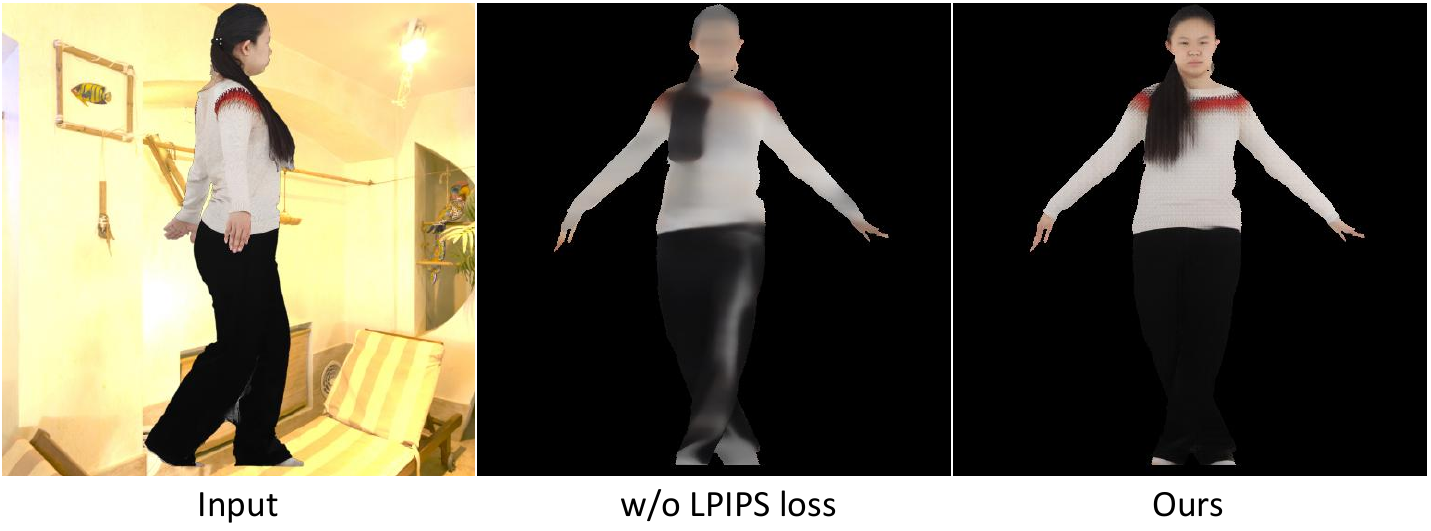}
    \captionsetup{type=figure}
    \captionof{figure}{Ablation study comparing our model with and \yl{without} the LPIPS loss.}
    \label{fig:ablation_loss}
    \vspace{-0.3cm}
\end{figure}

\subsection{Ablation study}
We conducted ablation experiments to compare variants of our architecture, different training strategies and different testing strategies.

\noindent\textbf{Effects of Fourier occupation field (FOF) in the encoder.} We remove the FOF \yl{input} during encoding to verify the effects of 3D information on the results. \yl{Quantitative} comparisons are in the sixth row of Tab.~\ref{information}, showing that all three metrics of rendering results without FOF drop. As shown in Fig. \ref{fig:ablation_zmap}, FOF can help image encoder better perceive 3D information, thereby reducing rendering artifacts and \yl{blurriness}.

\yyw{\noindent\textbf{Effects of Z-map in the \yyw{decoder}.} We retrained a variant without Z-map to explore the impact of Z-map. The first line of Tab.~\ref{information} shows the quantitative comparison results, and Fig.~\ref{fig:ablation_zmap} shows the qualitative comparison results. It can be seen that Z-map can effectively improve the details of the generated results.}

\yyw{\noindent\textbf{Effects of normal map in the rendering network.} 
Fig.~\ref{fig:ablation_normal} illustrates the effects of incorporating normal maps into the rendering network. Normal maps are not generated by neural networks but are rendered from the reconstructed mesh. The second line of Tab.~\ref{information} shows the quantitative results. Normal maps can provide the network with more surface information, which allows the network to better obtain the correct color from the texture field and reduce the generation of artifacts.}

\yyw{\noindent\textbf{Effects of FOF in the \yyw{encoder}.} Fig.~\ref{fig:ablation_fof} shows the effect of using FOF in the encoder on the results, and the third line of Tab.~\ref{information} shows the quantitative results. FOF can help the encoder better understand the spatial information of the human mesh, thus enabling the network to synthesize accurate colors for large invisible areas, when the viewing angle changes substantially.}

\begin{table}[htb]
    \centering
    \caption{Quantitative evaluation for ablation study on THuman2.0 dataset.}
    \label{information}
    \resizebox{.4\textwidth}{!}{%
    \begin{tabular}{c|ccc}
    \toprule[1.5pt]
    Method &SSIM$\uparrow$ &PSNR$\uparrow$ &LPIPS$\downarrow$\\ 
    \hline
    w/o $Z_{map}$   &0.9570 &30.18  &0.0392\\
    w/o normal      &\textbf{0.9571} &30.26  &0.0390\\
    w/o FOF         &0.9567 &29.93  &0.0394\\
    w/o LPIPS loss    &0.9382 &25.81  &0.0808\\
    \hline
    Ours(full) &\textbf{0.9571}&\textbf{30.52}&\textbf{0.0387}\\
    \bottomrule[1.5pt]
    \end{tabular}%
    }
\end{table}

\noindent\textbf{Effects of LPIPS loss.} To explore the effects of perceived loss on training, we retrained a variant without using \yl{the} perceived loss. Fig.~\ref{fig:ablation_loss} compares results obtained with and without the use of perceived loss, and we also make a quantitative comparison in the \yyw{fourth} line of Tab.~\ref{information}. The results show that the LPIPS loss \yyw{greatly improves the network's ability to learn the texture information, resulting in photo-realistic rendering results.}

\begin{figure*}[htb]
    \centering
    \includegraphics[width=.9\textwidth]{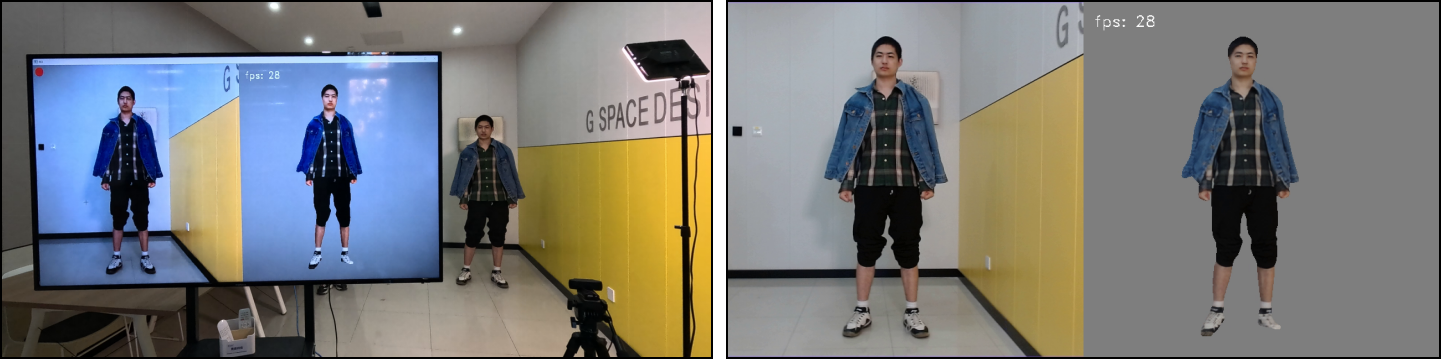}
    \captionsetup{type=figure}
    \captionof{figure}{Real-time rendering results by our method.}
    \label{fig:realtime}
\end{figure*}
\begin{figure*}[htb]
    \centering
    \includegraphics[width=.9\textwidth]{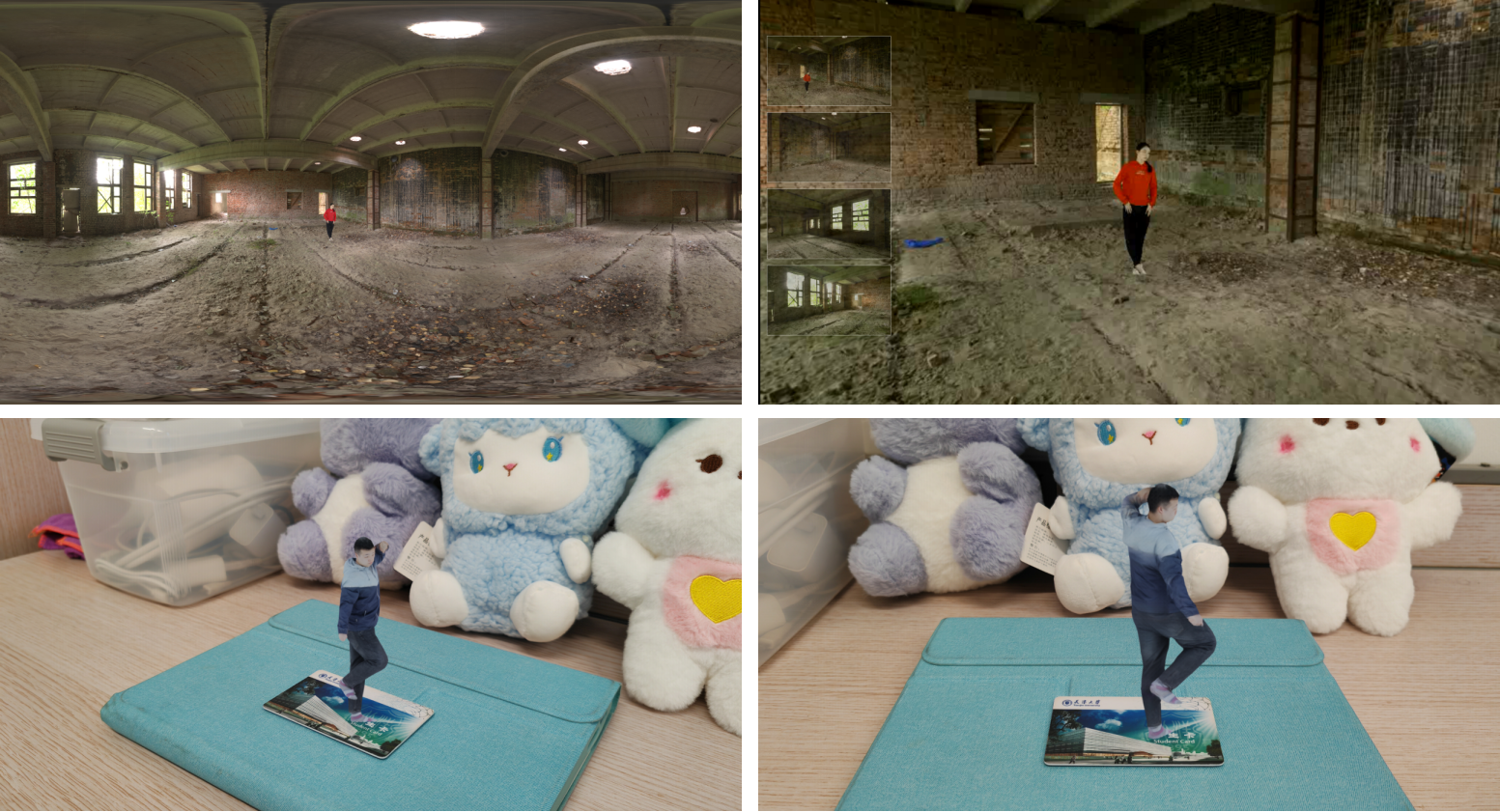}
    \captionsetup{type=figure}
    \captionof{figure}{VR/AR rendering results by our method.}
    \label{fig:vr}
\end{figure*}

\vspace{+0.125cm}
\yyw{\noindent\textbf{Effects of consistency strategy.} We add a consistency loss and a spatial fusion strategy to ensure the multi-view consistency of the results. To explore the impact of the spatial fusion strategy on the results, we retrain a variant without the consistency loss and use the method described in~\cite{vianello2018robust} to draw EPI images to qualitatively evaluate the view consistency. Fig. \ref{fig:ablation_epi} compares the results with and without the spatial fusion strategy, and it can be seen that our spatial fusion strategy effectively enhances the view consistency during rotation.}

\subsection{Real-time Rendering System}
Our pipeline, designed for real-time monocular human novel view synthesis, operates as follows: Images are captured from the video stream using OpenCV~\cite{bradski2000opencv} and preprocessed to a resolution of $512 \times 512$. We then apply FOF-SMPL~\cite{feng2022fof} to infer the occupation field. Subsequently, we employ the marching cube algorithm to extract the mesh and render the depth and normal maps. The final high-fidelity novel view images are rendered in real-time with the camera parameters $\Pi^{t}$ and the previously processed data. Our pipeline, implemented with TensorRT on a single RTX-3090 GPU, achieves a performance of 24+ FPS, with the potential for further enhancement via additional GPUs.
Fig. \ref{fig:realtime} showcase some of the real-time reconstruction results. Although our method trains to process each frame separately using synthetic data, it can still provide reasonable temporal consistency. Please refer to the demo video for more results.

\subsection{Virtual Reality and Augmented Reality Applications}
\yywnew{Our method has many meaningful applications, and we present two applications here.} As shown in the top row of Fig. \ref{fig:vr}, we render the human appearance into a panoramic photo to generate the person in a virtual environment and display it in VR glasses. Specifically, we project the results into a spherical coordinate system and expand the human appearance synthesized into the panoramic image to obtain the panoramic human appearance. By rendering the 3D human appearance into the virtual environment, our method  provides people with a more realistic and stunning virtual experience.
\yywnew{In addition, our method can also be integrated into the real scene of AR applications. As shown in the bottom row of Fig. \ref{fig:vr}, given the camera parameters of each frame in the video, we can render the human appearance in a specific scene and observe the 3D human from different views. Our VR/AR examples can generate many realistic applications in certain situations, \eg, AR/VR education, immersion effects in games, virtual character communication in real scenes, etc. Also, our method of realistic 3D human appearance rendering has promising applications in immersive telepresence. In virtual conferences, the real 3D human appearance can greatly increase people's immersion. Applying our method in these VR/AR applications is a valuable direction in the future.}

\section{Conclusion and Discussion}
\noindent\textbf{Conclusion.} Rendering 3D human appearance from a single image in real-time is important for achieving holographic communication and immersive
VR/AR. We propose a novel method, combining the strengths of implicit texture field and explicit neural rendering, for real-time inference and rendering of realistic 3D human appearance from a monocular RGB image. We propose a new representation, Z-map,  to alleviate depth ambiguities in rendering and enable high-fidelity color reconstruction. We also design a consistency loss and a spatial fusion strategy to ensure the multi-view coherence and reduce the jittering phenomenon. Experimental results show that the proposed method achieves state-of-the-art performance on both synthetic data and real-world images.


\noindent\textbf{Broader Impact.} Our proposed framework has the potential to significantly advance the development of holographic communication. However, high-fidelity novel view synthesis may also raise privacy concerns. Therefore, we strongly recommend that regulators establish ethical guidelines and regulatory frameworks that strike a balance between innovation and privacy protection.





\acknowledgments{
This work was supported in part by the National Natural Science Foundation of China (62122058 and 62171317), and the Science Fund for Distinguished Young Scholars of Tianjin (No. 22JCJQJC00040).}

\bibliographystyle{abbrv-doi}

\bibliography{template}
\clearpage

\end{document}